\documentclass[sn-mathphys-num]{sn-jnl}

\usepackage{graphicx}%
\usepackage{multirow}%
\usepackage{amsmath,amssymb,amsfonts}%
\usepackage{amsthm}%
\usepackage{mathrsfs}%
\usepackage[title]{appendix}%
\usepackage{xcolor}%
\usepackage{textcomp}%
\usepackage{manyfoot}%
\usepackage{booktabs}%
\usepackage{algorithm}%
\usepackage{algorithmicx}%
\usepackage{algpseudocode}%
\usepackage{listings}%
\usepackage{graphicx}
\usepackage{amsmath}
\usepackage{amsthm}
\usepackage{booktabs}
\usepackage[switch]{lineno}

\usepackage{xcolor}
\usepackage{graphicx}
\usepackage{multirow}
\usepackage{booktabs}
\usepackage{tabularx}
\usepackage{tikz}
\usepackage{hyperref}
\newcolumntype{I}{!{\vrule width 2pt}}
\usetikzlibrary{decorations.pathreplacing,positioning, arrows.meta}

\theoremstyle{thmstyleone}%
\theoremstyle{thmstyletwo}%

\theoremstyle{thmstylethree}%

\begin{document}

\title[Article Title]{Survey and Taxonomy: The Role of Data-Centric AI in Transformer-Based Time Series Forecasting}


\author[1]{\fnm{Jingjing} \sur{Xu}}

\author[2]{\fnm{Caesar} \sur{Wu}}

\author[3]{\fnm{Yuan-Fang} \sur{Li}}

\author[1,2]{\fnm{Grégoire} \sur{Danoy}}

\author[1,2]{\fnm{Pascal} \sur{Bouvry}}

\affil[1]{\orgdiv{FSTM/DCS}, \orgname{University of Luxembourg}}

\affil[2]{\orgdiv{SnT}, \orgname{University of Luxembourg}
}

\affil[3]{\orgdiv{Faculty of Information Technology}, \orgname{Monash University}}


\abstract{
Alongside the continuous process of improving AI performance through the development of more sophisticated models, researchers have also focused their attention to the emerging concept of data-centric AI, which emphasizes the important role of data in a systematic machine learning training process. Nonetheless, the development of models has also continued apace. One result of this progress is the development of the Transformer Architecture, which possesses a high level of capability in multiple domains such as Natural Language Processing (NLP), Computer Vision (CV) and Time Series Forecasting (TSF). Its performance is, however, heavily dependent on input data preprocessing and output data evaluation, justifying a data-centric approach to future research. We argue that data-centric AI is essential for training AI models, particularly for transformer-based TSF models efficiently. However, there is a gap regarding the integration of transformer-based TSF and data-centric AI. This survey aims to pin down this gap via the extensive literature review based on the proposed taxonomy. We review the previous research works from a data-centric AI perspective and we intend to lay the foundation work for the future development of transformer-based architecture and data-centric AI.
~\footnote{This work was funded by the Luxembourg National Research Fund (Fonds National de la Recherche - FNR), Grant ID 15748747 and Grant ID C21/IS/16221483/CBD. We thank to Dr. Pierrick Pochelu from University of Luxembourg and Jean-Francois Nies from Luxembourg Institute of Science and Technology (LIST) for proofreading. We thank to Lucas Villiere from University of Luxembourg for measuring the carbon footprint of the models. Also, we thank to HPC group of the University of Luxembourg}
}
\keywords{Transformer, Time Series Forecasting, Data-Centric AI}



\maketitle

\section{Introduction}
Sequential data like languages and time series can be effectively modeled using recurrent neural networks (RNNs)~\cite{schuster1997bidirectional}, which connect features across different points in a sequence. Traditional fully connected neural networks (FCNs) lack this capability. However, RNNs face challenges such as vanishing and exploding gradients, which impede their ability to capture long-term dependencies. Solutions include using Rectified Linear Unit (ReLU) activation functions, initializing weights with identity matrices, and implementing gates to control information flow. Architectures like Gated Recurrent Units (GRUs)~\cite{cho2014learning} and Long Short-Term Memory (LSTM)~\cite{hochreiter1997long} networks use gate cells to manage long-term dependencies, but some challenges remain, particularly in training, which often considers only one direction of data leading to capture less detailed context. Bidirectional RNNs address this by processing data in both forward and reverse directions. Despite improvements, these models lack parallelism, leading to inefficiencies.

The Transformer~\cite{vaswani2017attention} architecture offers an attractive solution to these issues by processing sequence data in parallel through an encoder-decoder architecture and applying multi-head attention mechanisms, thereby significantly enhancing efficiency and performance. Transformers have proven to be highly effective in processing long-term sequence data, such as lengthy sentences and speech in NLP~\cite{gillioz2020overview} applications, as well as handling image and video data in CV applications~\cite{han2022survey,nies2024vision}. Their capabilities in the realm of long time series forecasting have also been recognized.

\begin{figure}[ht]
\begin{center}
\includegraphics[clip, trim=2cm 0cm 1cm 0cm, width=0.8\linewidth]{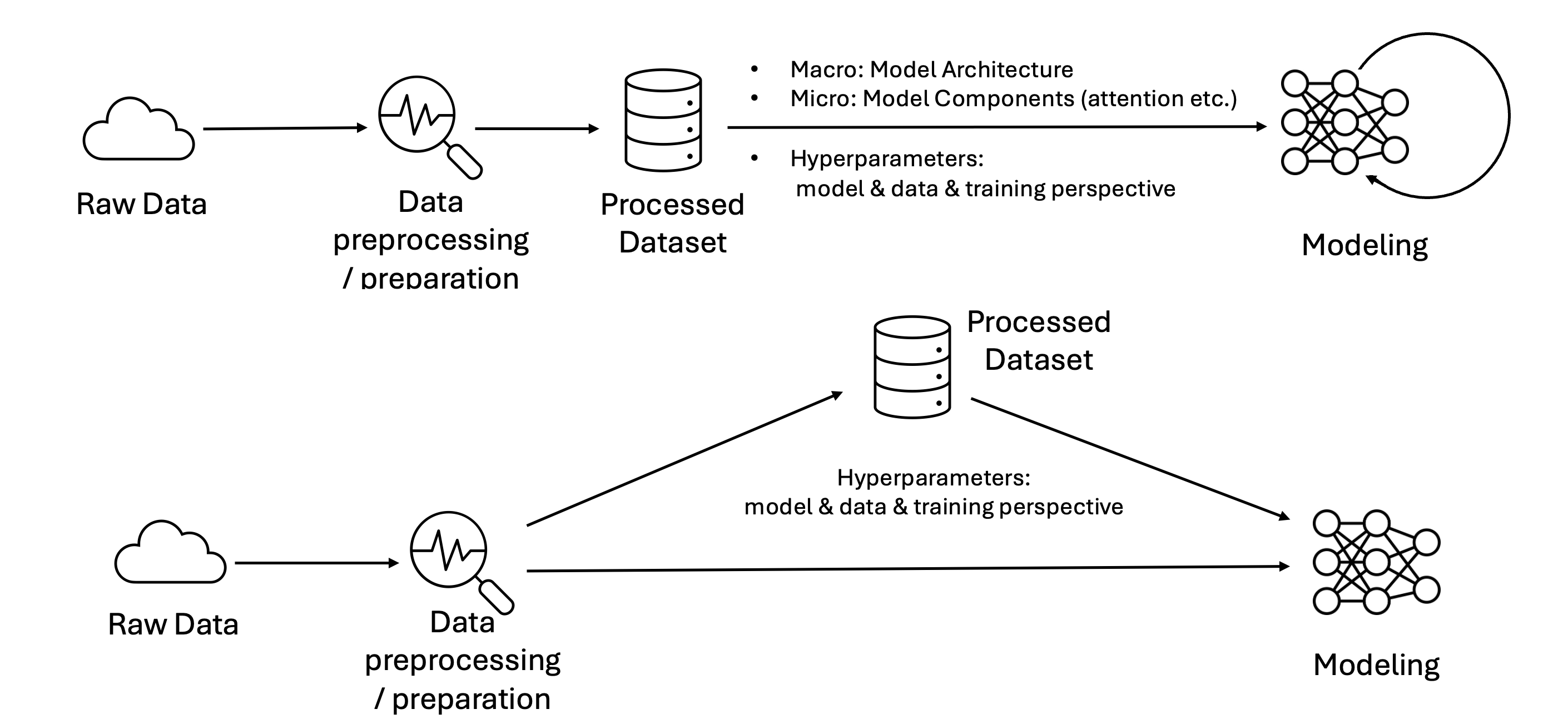}
\end{center}
\caption{MCAI (top) and DCAI (bottom) Iteration Life Cycles.}
\label{MCAI_L}
\end{figure}
Nevertheless, the improvements made by these model-centric AI approaches (MCAI) can not always overcome insufficient data preparation and preprocessing. These steps, which are essential for developing precise models, were often neglected. Therefore, researchers are increasingly directing their efforts towards data-centric approaches to achieve better model performance, leading to the emergence of Data-Centric AI (DCAI)~\cite{polyzotis2021can,jakubik2022data,jarrahi2022principles,whang2023data,zha2023data1,zha2023data2}. According to Andrew Ng, data-centric AI is the discipline of systematically engineering the data used to build an AI system~\cite{data2022data}. It emphasizes the importance of data in the analysis compared to conventional model-centric AI. We illustrate the processes of data-centric AI (DCAI) and model-centric AI (MCAI) in Fig.~\ref{MCAI_L}. However, existing surveys and tutorials focus on explaining the role of MCAI in transformer-based time series models.Consequently, there is a lack of surveys on DCAI in transformer-based time series models. (N.B. In this paper, the terms transformer architecture and transformer model are used interchangeably.) 
\begin{figure}
\begin{center}
\includegraphics[width=0.5\linewidth]{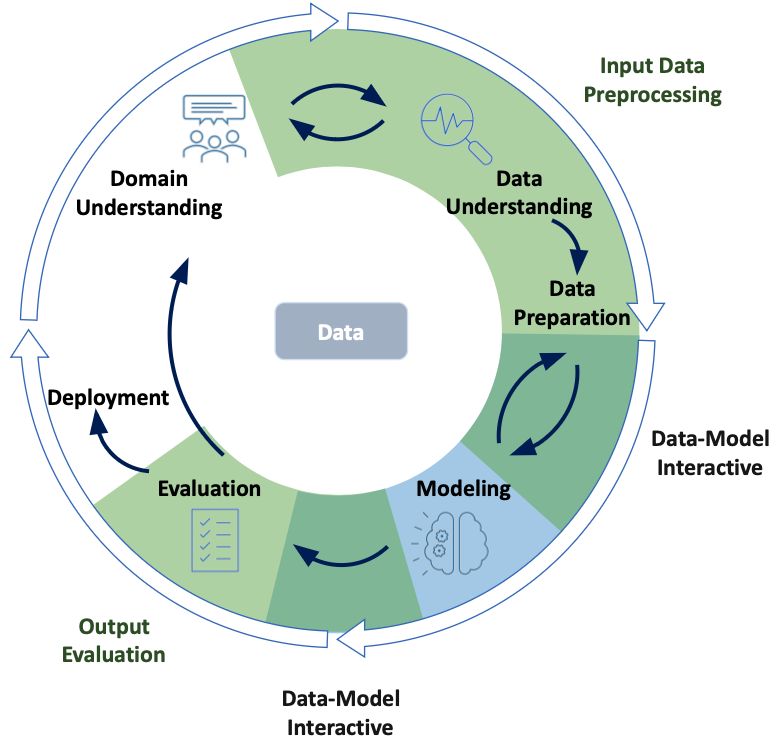}
\end{center}
\caption{Different phases of CRISP-DM in Data-Centric AI perspective.}
\label{CRISP-DM-AI}
\end{figure}

To address this gap, we apply the CRoss-Industry Standard Process for Data Mining (CRISP-DM) (see Fig.~\ref{CRISP-DM-AI}) to build a taxonomy (Sec.~\ref{taxonomy}) for data-centric time series forecasting. CRISP-DM is viewed as an AI system-building process, dividing the process into six major phases: domain (business) understanding, data understanding, data preparation (preprocessing), modeling, evaluation, and deployment~\cite{wirth2000crisp}. We categorize these processes into Input data preprocessing, data-model interaction, and output evaluation in our taxonomy (See Fig.~\ref{taxonomytree}). Our contributions encompass several key aspects. We propose a new taxonomy based on data-centric AI and address the following Research Questions (RQs):
{\small
\begin{quote}
{\bf RQ 1: }\textit{
How are datasets preprocessed before being fed into models?}\\
{\bf RQ 2: }\textit{
How does the data and model interact with each other?}\\
{\bf RQ 3: }\textit{
How are models evaluated on the data?} \\
\end{quote}
}
We propose {\bf RQ1} because the initial step of constructing a model involves a thorough analysis of the time series data~\cite{hyndman2018forecasting,abedjan2017data}. Without this fundamental understanding, the justification for model construction becomes uncertain, as models typically do not perform well with random and unstructured datasets. Furthermore, applying the model effectively and efficiently with given datasets in practical forecasting tasks is challenging. Therefore, addressing {\bf RQ2} is crucial for building an effective and efficient model. To successfully implement models in practical forecasting applications, it is necessary to evaluate their performance, making {\bf RQ3} vital. The paper is organized as follows: Section~\ref{re_surveys} describes related surveys. Section~\ref{taxonomy} provides the taxonomy. Section~\ref{datasets} addresses RQ1, Section~\ref{interaction} addresses RQ2, and Section~\ref{evaluation} addresses RQ3. Section~\ref{challenges} indicate future research opportunities. Section~\ref{Conclusion} concludes the paper.




\section{Related Works and Background}
\label{re_surveys}
\subsection{Data-Centric Artificial Intelligence (Data-Centric AI)}

With the development and movement~\cite{ng2021data,ng2021chat} of Data-Centric AI (DCAI), several surveys have emerged on this topic. However, none have specifically addressed the role of DCAI in Transformer models for time series forecasting. Some researchers~\cite{jarrahi2022principles} have introduced the concepts and principles to outline the foundations of DCAI. Others~\cite{jakubik2022data} have defined relevant terms and introduced a framework for DCAI, while another study~\cite{zha2023data1} provides a overview of DCAI missions, detailing definitions, explanations, related tasks, and challenges. These surveys primarily focus on clarifying the definitions and frameworks or providing an overview of DCAI, without delving into detailed methods for each section. One comprehensive survey~\cite{zha2023data2} examines the entire data lifecycle with representative methods but does not discuss specific use cases. Another survey~\cite{yang2023data} explores use cases with graph learning, and another~\cite{rodriguez2022data} provides an epidemic forecasting survey from a DCAI perspective. However, none of these involve Transformer models. One paper~\cite{ivanov2021data} studies data movement in Transformers' training processes to improve GPU utilization, but it does not consider the data itself. Another study~\cite{dan2023transface} addresses the performance issues of Vision Transformers (ViTs) in face recognition scenarios from a data-centric perspective. Nevertheless, the role of DCAI in Transformer-based time series forecasting remains insufficiently explored.

\subsection{Transformer for Time Series Forecasting}
\subsubsection{Time Series Forecasting}
Surveys~\cite{lim2021time} and tutorials~\cite{benidis2022deep} discuss deep learning for time series forecasting from the perspective of model architectures, while another review~\cite{lara2021experimental} conducts experimental studies to compare the performance of different deep learning architectures. The Monash time series forecasting archive~\cite{godahewa2021monash} provides a diverse collection of comprehensive time-series datasets across various domains, along with dataset characteristics analysis. However, these surveys do not conduct comprehensive analyses of Transformers in time series forecasting, and their dataset analysis and analysis pipelines are deficient.
\subsubsection{Transformer for Time Series} 
The use of Transformers for time series tasks arises from their powerful capability in handling sequential data. A survey~\cite{wen2023transformers} analyzes the development of time series Transformers from network modifications and application domain perspectives. A tutorial~\cite{ahmed2023transformers} provides details about Transformer architecture and its applications. However, these studies approach the topic from a model-centric AI perspective, leaving the survey of Transformers in time series from a data-centric AI perspective insufficient. Thus, we introduce the role of DCAI in Transformer-based time series forecasting to fill this gap. We particularly focus on explaining data-model interaction in Transformer architecture. Details on this interaction in time series are introduced in Sec.~\ref{interaction}, covering input time series data representation part and the modeling part. Transformer data representation includes input embedding and position encoding, crucial for time series due to the importance of position order. A review~\cite{dufter2022position} provides an overview of position information in Transformers to help choose suitable position encoding solutions. Another paper~\cite{foumani2024improving} studies different position encoding and proposes a solution for multivariate time series classification called ConvTran. Nevertheless, the study of input embedding and position encoding for time series forecasting still requires further exploration.
\begin{figure*}[!ht]
\begin{center}
\includegraphics[width=\linewidth]{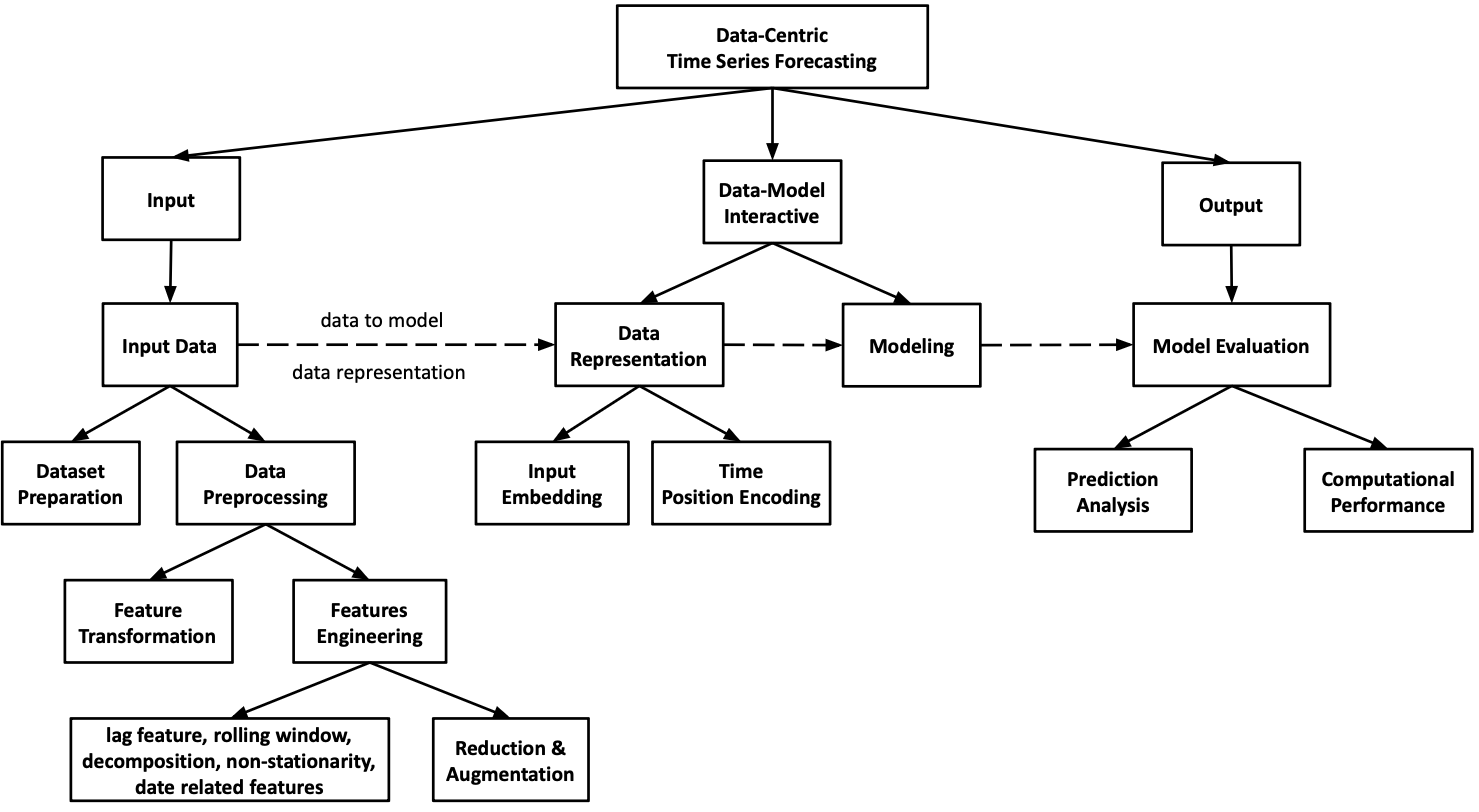}
\end{center}
\caption{Taxonomy of Data-Centric Transformer-based Time Series Forecasting}
\label{taxonomytree}
\end{figure*}




\section{Taxonomy}
\label{taxonomy}
We adopt the CRISP-DM and Transformer model workflow as the foundational framework to systematically structure the different phases of our survey, as illustrated in Figure~\ref{taxonomytree}. Input Data, Data-Model Interaction, and Output Evaluation will be detailed in Sections~\ref{datasets},~\ref{interaction}, and~\ref{evaluation} respectively. In the Input Data section, we cover data preparation and preprocessing to address RQ1: \textit{How are datasets preprocessed before being fed into models?} In the Data-Model Interaction section, we discuss embedding, encoding, and modeling to address RQ2: \textit{How does the data and model interact with each other?} In the Output Evaluation section, we explain model evaluation to address RQ3: \textit{How are models evaluated on the data?}

\begin{table*}[!ht]
    \caption{Time Series Forecasting Models and Datasets (Open-Source Transformer model and LLM until 2023. Nov. 01)}
    \label{Tran_LM}
    \centering
    \setlength\aboverulesep{0pt} \setlength\belowrulesep{0pt}
    \resizebox{\linewidth}{!}{
    \begin{tabular}{c|cIc|c|c|c|c|c|c|c|c|c|c|c|c|c|c|c|c|c|c|cIc|c|c|c|c}
    \toprule[2pt]
        ~ & Model & \multicolumn{20}{cI}{Transformer} & \multicolumn{4}{c|}{LLM} & ~\\ 
        \cmidrule{2-26}
        \rotatebox{90}{Domain} & Abbreviation & \rotatebox{90}{GCformer~\cite{zhao2023gcformer}}  & \rotatebox{90}{PDTrans~\cite{tong2023probabilistic}} & \rotatebox{90}{PatchTST~\cite{nie2022time}} & \rotatebox{90}{Crossformer~\cite{zhang2022crossformer}} & \rotatebox{90}{Scaleformer~\cite{shabani2022scaleformer}} & \rotatebox{90}{Preformer~\cite{du2023preformer}} & \rotatebox{90}{Client~\cite{gao2023client}} & \rotatebox{90}{Taylorformer~\cite{nivron2023taylorformer}} & \rotatebox{90}{iTransformer~\cite{liu2023itransformer}} & \rotatebox{90}{TDformer~\cite{zhang2022first}} & \rotatebox{90}{Non-stationary Trans.~\cite{liu2022non}} & \rotatebox{90}{FEDformer~\cite{zhou2022fedformer}} & \rotatebox{90}{TACTiS~\cite{drouin2022tactis}} & \rotatebox{90}{Pyraformer~\cite{liu2021pyraformer}} & \rotatebox{90}{TCCT~\cite{shen2022tcct}} & \rotatebox{90}{Triformer~\cite{cirsteatriformer}} & \rotatebox{90}{Autoformer~\cite{wu2021autoformer}} & \rotatebox{90}{Informer~\cite{zhou2021informer}} & \rotatebox{90}{TFT~\cite{lim2021temporal}} & \rotatebox{90}{AST~\cite{wu2020adversarial}} & \rotatebox{90}{PromptCast~\cite{xue2023promptcast}} & \rotatebox{90}{One Fits All~\cite{zhou2023one}} & \rotatebox{90}{LLMTime~\cite{gruver2024large}} & \rotatebox{90}{Lag-Llama~\cite{rasul2024lagllama}} & \rotatebox{90}{Frequency} \\ 
        \cmidrule{2-26}
        ~ & Implement & \href{https://github.com/Yanjun-Zhao/Gcformer}{code} & \href{https://github.com/JL-tong/PDTrans}{code} & \href{https://github.com/yuqinie98/PatchTST}{code} & \href{https://github.com/Thinklab-SJTU/Crossformer}{code} & \href{https://github.com/BorealisAI/scaleformer}{code} & \href{https://github.com/ddz16/Preformer}{code} & \href{https://github.com/daxin007/client}{code} & \href{https://www.dropbox.com/s/vnxuwq7zm7m9bj8/taylorformer.zip?dl=0}{code} & \href{https://github.com/thuml/iTransformer}{code} & \href{https://github.com/BeBeYourLove/TDformer}{code} & \href{https://github.com/thuml/Nonstationary_Transformers}{code} & \href{https://github.com/DAMO-DI-ML/ICML2022-FEDformer}{code} & \href{https://github.com/ServiceNow/tactis}{code} & \href{https://github.com/alipay/Pyraformer}{code} & \href{https://github.com/OrigamiSL/TCCT2021}{code} & \href{https://github.com/razvanc92/triformer}{code} & \href{https://github.com/thuml/autoformer}{code} & \href{https://github.com/zhouhaoyi/Informer2020}{code} & \href{https://github.com/google-research/google-research/tree/master/tft}{code} & \href{https://github.com/hihihihiwsf/AST}{code} & \href{https://github.com/HaoUNSW/PISA}{code} & \href{https://github.com/DAMO-DI-ML/NeurIPS2023-One-Fits-All}{code} & \href{https://github.com/ngruver/llmtime}{code} & \href{https://github.com/time-series-foundation-models/lag-llama}{code} & ~ \\ 
        \midrule[1.2pt]
        Energy & \href{https://zenodo.org/records/4656140}{Electricity} & $\surd$ & $\surd$ & $\surd$ & $\surd$ & $\surd$ & $\surd$ & $\surd$ & $\surd$ & $\surd$ & $\surd$ & $\surd$ & $\surd$ & $\surd$ & $\surd$ & $\surd$ & $\surd$ & $\surd$ & $\surd$ & $\surd$ & $\surd$ & $\surd$ & $\surd$ & $\surd$ & $\surd$ & 24 \\ 
        \hline
        Transport & \href{https://zenodo.org/records/4656132}{Traffic Hourly} & $\surd$ & $\surd$ & $\surd$ & $\surd$ & $\surd$ & $\surd$ & $\surd$ & ~ & $\surd$ & $\surd$ & $\surd$ & $\surd$ & $\surd$ & ~ & $\surd$ & ~ & $\surd$ & ~ & $\surd$ & $\surd$ & ~ & $\surd$ & $\surd$ & ~ & 18 \\ 
        \hline
        Finance  & \href{https://github.com/laiguokun/multivariate-time-series-data}{Exchange} & $\surd$ & $\surd$ & ~ & ~ & $\surd$ & $\surd$ & $\surd$ & $\surd$ & ~ & $\surd$ & $\surd$ & $\surd$ & ~ & ~ & ~ & ~ & $\surd$ & ~ & ~ & ~ & ~ & ~ & $\surd$ & $\surd$ & 12 \\ 
        \hline
        Energy & \href{https://github.com/zhouhaoyi/ETDataset}{ETTm1} & $\surd$ & ~ & $\surd$ & $\surd$ & ~ & ~ & $\surd$ & $\surd$ & ~ & $\surd$ & ~ & ~ & ~ & $\surd$ & $\surd$ & $\surd$ & $\surd$ & $\surd$ & ~ & ~ & ~ & $\surd$ & ~ & ~ & 12 \\ 
        \hline
        Nature & \href{https://drive.google.com/drive/folders/1ZOYpTUa82_jCcxIdTmyr0LXQfvaM9vIy}{Weather\_DE} & $\surd$ & ~ & $\surd$ & ~ & $\surd$ & $\surd$ & $\surd$ & ~ & ~ & $\surd$ & $\surd$ & $\surd$ & ~ & ~ & ~ & ~ & $\surd$ & ~ & ~ & ~ & ~ & $\surd$ & $\surd$ & ~ & 11 \\ 
        \hline
        Energy & \href{https://github.com/zhouhaoyi/ETDataset}{ETTh1} & ~ & ~ & $\surd$ & $\surd$ & ~ & ~ & $\surd$ & ~ & $\surd$ & ~ & ~ & ~ & ~ & $\surd$ & $\surd$ & $\surd$ & $\surd$ & $\surd$ & ~ & ~ & ~ & $\surd$ & ~ & ~ & 10 \\ 
        \hline
        Energy & \href{https://github.com/zhouhaoyi/ETDataset}{ETTm2} & $\surd$ & ~ & $\surd$ & ~ & ~ & $\surd$ & $\surd$ & ~ & ~ & ~ & $\surd$ & $\surd$ & ~ & ~ & $\surd$ & ~ & $\surd$ & ~ & ~ & ~  & ~ & $\surd$ & $\surd$ & ~ & 10 \\ 
        \hline
        Health & \href{https://drive.google.com/drive/folders/1DasX30lzEwcVXYaNeyMlQ0PSmCQSow5h}{ILI} & $\surd$ & ~ & $\surd$ & $\surd$ & $\surd$ & $\surd$ & $\surd$ & ~ & ~ & ~ & $\surd$ & $\surd$ & ~ & ~ & ~ & ~ & $\surd$ & ~ & ~ & ~  & ~ & $\surd$ & ~ & ~ & 10 \\ 
        \hline
        Energy & \href{https://github.com/zhouhaoyi/ETDataset}{ETTh2}  & ~ & ~ & $\surd$ & ~ & ~ & ~ & $\surd$ & ~ & $\surd$ & ~ & ~ & ~ & ~ & ~ & $\surd$ & ~ & $\surd$ & $\surd$ & ~ & ~ & ~ & $\surd$ & ~ & ~ & 7 \\ 
        \hline
        Energy & \href{https://zenodo.org/records/4656144}{Solar 10 minutes} & ~ & $\surd$ & ~ & ~ & ~ & ~ & ~ & ~ & $\surd$ & ~ & ~ & ~ & $\surd$ & ~ & ~ & ~ & ~ & ~ & ~ & $\surd$ & ~ & ~ & $\surd$ & $\surd$ & 6 \\ 
        \hline
        Nature & \href{https://www.ncei.noaa.gov/data/local-climatological-data/}{Weather\_US} & ~ & ~ & ~ & $\surd$ & ~ & ~ & ~ & ~ & ~ & ~ & ~ & ~ & ~ & ~ & ~ & $\surd$ & ~ & $\surd$ & ~ & ~ & ~ & ~ & ~ & ~ & 3 \\ 
        \hline
        Finance  & \href{https://zenodo.org/records/4654833}{Fred-MD} & ~ & ~ & ~ & ~ & ~ & ~ & ~ & ~ & ~ & ~ & ~ & ~ & $\surd$ & ~ & ~ & ~ & ~ & ~ & ~ & ~ & ~ & ~ & $\surd$ & $\surd$ & 3 \\ 
        \hline
        Multiple & \href{https://zenodo.org/records/4656589}{M4 Hourly} & ~ & $\surd$ & ~ & ~ & ~ & ~ & ~ & ~ & ~ & ~ & ~ & ~ & ~ & ~ & ~ & ~ & ~ & ~ & ~ & $\surd$ & ~ & ~ & ~ & $\surd$ & 3 \\ 
        \hline
        Nature & \href{https://www.kaggle.com/datasets/sohier/30-years-of-european-wind-generation}{Wind\_EU} & ~ & ~ & ~ & ~ & ~ & ~ & ~ & ~ & ~ & ~ & ~ & ~ & ~ & $\surd$ & ~ & ~ & ~ & ~ & ~ & $\surd$ & ~ & ~ & ~ & ~ & 2 \\ 
        \hline
        Health & \href{https://zenodo.org/records/4656009}{Covid Deaths} & ~ & ~ & ~ & ~ & ~ & ~ & ~ & ~ & ~ & ~ & ~ & ~ & ~ & ~ & ~ & ~ & ~ & ~ & ~ & ~ & ~ & ~ & $\surd$ & $\surd$ & 2 \\ 
        \hline
        Tourism & \href{https://zenodo.org/records/4656096}{Tourism Monthly} & ~ & ~ & ~ & ~ & ~ & ~ & ~ & ~ & ~ & ~ & ~ & ~ & ~ & ~ & ~ & ~ & ~ & ~ & ~ & ~ & ~ & ~ & $\surd$ & $\surd$ & 2 \\ 
        \hline
        Energy & \href{https://zenodo.org/records/4659727}{Aus. Electricity}  & ~ & ~ & ~ & ~ & ~ & ~ & ~ & ~ & ~ & ~ & ~ & ~ & ~ & ~ & ~ & ~ & ~ & ~ & ~ & ~ & ~ & ~ & $\surd$ & $\surd$ & 2 \\ 
        \hline
        Transport & \href{https://zenodo.org/records/4656626}{Pedestrian Counts} & ~ & ~ & ~ & ~ & ~ & ~ & ~ & ~ & ~ & ~ & ~ & ~ & ~ & ~ & ~ & ~ & ~ & ~ & ~ & ~ & ~ & ~ & $\surd$ & $\surd$ & 2 \\ 
        \hline
        Health & \href{https://zenodo.org/records/4656014}{Hospital} & ~ & ~ & ~ & ~ & ~ & ~ & ~ & ~ & ~ & ~ & ~ & ~ & ~ & ~ & ~ & ~ & ~ & ~ & ~ & ~ & ~ & ~ & $\surd$ & $\surd$ & 2 \\
        \hline
        Banking & \href{https://zenodo.org/records/4656125}{NN5 weekly} & ~ & ~ & ~ & ~ & ~ & ~ & ~ & ~ & ~ & ~ & ~ & ~ & ~ & ~ & ~ & ~ & ~ & ~ & ~ & ~ & ~ & ~ & $\surd$ & $\surd$ & 2 \\ 
        \hline
        Nature & \href{https://zenodo.org/records/4656058}{Saugeen River Flow} & ~ & ~ & ~ & ~ & ~ & ~ & ~ & ~ & ~ & ~ & ~ & ~ & ~ & ~ & ~ & ~ & ~ & ~ & ~ & ~ & ~ & ~ & $\surd$ & $\surd$ & 2 \\ 
        \hline
        Finance & \href{https://zenodo.org/records/4656042}{CIF 2016} & ~ & ~ & ~ & ~ & ~ & ~ & ~ & ~ & ~ & ~ & ~ & ~ & ~ & ~ & ~ & ~ & ~ & ~ & ~ & ~ & ~ & ~ & $\surd$ & $\surd$ & 2 \\ 
        \hline
        Nature & \href{https://zenodo.org/records/4656756}{KDD Cup 2018} & ~ & ~ & ~ & ~ & ~ & ~ & ~ & ~ & ~ & ~ & ~ & ~ & $\surd$ & ~ & ~ & ~ & ~ & ~ & ~ & ~ & ~ & ~ & ~ & $\surd$ & 2 \\ 
        \hline
        Transport & \href{https://github.com/thuml/iTransformer}{PEMS} & ~ & ~ & ~ & ~ & ~ & ~ & ~ & ~ & $\surd$ & ~ & ~ & ~ & ~ & ~ & ~ & ~ & ~ & ~ & ~ & ~ & ~ & ~ & ~ & ~ & 1 \\ 
        \hline
        Sales & \href{https://github.com/google-research/google-research/blob/master/tft/script_download_data.py}{Retail} & ~ & ~ & ~ & ~ & ~ & ~ & ~ & ~ & ~ & ~ & ~ & ~ & ~ & ~ & ~ & ~ & ~ & ~ & $\surd$ & ~ & ~ & ~ & ~ & ~ & 1 \\ 
        \hline
        Finance & \href{https://github.com/onnokleen/mfGARCH/tree/v0.1.9/data-raw}{Volatility} & ~ & ~ & ~ & ~ & ~ & ~ & ~ & ~ & ~ & ~ & ~ & ~ & ~ & ~ & ~ & ~ & ~ & ~ & $\surd$ & ~ & ~ & ~ & ~ & ~ & 1 \\ 
        \hline 
        Web & \href{https://github.com/ant-research/Pyraformer/tree/master/data}{App Flow}  & ~ & ~ & ~ & ~ & ~ & ~ & ~ & ~ & ~ & ~ & ~ & ~ & ~ & $\surd$ & ~ & ~ & ~ & ~ & ~ & ~ & ~ & ~ & ~ & ~ & 1 \\ 
        \hline
        Nature & \href{https://github.com/HaoUNSW/PISA}{City Temperature} & ~ & ~ & ~ & ~ & ~ & ~ & ~ & ~ & ~ & ~ & ~ & ~ & ~ & ~ & ~ & ~ & ~ & ~ & ~ & ~ & $\surd$ & ~ & ~ & ~ & 1 \\ 
        \hline
        Nature & \href{https://github.com/HaoUNSW/PISA}{SafeGraph Human Mobility} & ~ & ~ & ~ & ~ & ~ & ~ & ~ & ~ & ~ & ~ & ~ & ~ & ~ & ~ & ~ & ~ & ~  & ~ & ~ & ~ & $\surd$ & ~ & ~ & ~ & 1 \\ 
        \hline
        Energy & \href{https://zenodo.org/records/4656151}{Solar Weekly} & ~ & ~ & ~ & ~ & ~ & ~ & ~ & ~ & ~ & ~ & ~ & ~ & ~ & ~ & ~ & ~ & ~ & ~ & ~ & ~  & ~ & ~ & $\surd$ & ~ & 1 \\ 
        \hline
        Tourism & \href{https://zenodo.org/records/4656103}{Tourism Yearly} & ~ & ~ & ~ & ~ & ~ & ~ & ~ & ~ & ~ & ~ & ~ & ~ & ~ & ~ & ~ & ~ & ~ & ~ & ~ & ~ & ~ & ~ & $\surd$ & ~ & 1 \\ 
        \hline
        Tourism & \href{https://zenodo.org/records/4656093}{Tourism Quarterly} & ~ & ~ & ~ & ~ & ~ & ~ & ~ & ~ & ~ & ~ & ~ & ~ & ~ & ~ & ~ & ~ & ~ & ~ & ~ & ~ & ~ & ~ & $\surd$ & ~ & 1 \\ 
        \hline
        Nature & \href{https://zenodo.org/records/4656049}{US Births} & ~ & ~ & ~ & ~ & ~ & ~ & ~ & ~ & ~ & ~ & ~ & ~ & ~ & ~ & ~ & ~ & ~ & ~ & ~ & ~  & ~ & ~ & $\surd$ & ~ & 1 \\ 
        \hline
        Transport & \href{https://zenodo.org/records/4656135}{Traffic Weekly}  & ~ & ~ & ~ & ~ & ~ & ~ & ~ & ~ & ~ & ~ & ~ & ~ & ~ & ~ & ~ & ~ & ~ & ~  & ~ & ~ & ~ & ~ & $\surd$ & ~ & 1 \\ 
        \hline
        Sales & \href{https://zenodo.org/records/4656021}{Car Parts} & ~ & ~ & ~ & ~ & ~ & ~ & ~ & ~ & ~ & ~ & ~ & ~ & ~ & ~ & ~ & ~ & ~ & ~ & ~ & ~ & ~  & ~ & ~ & $\surd$ & 1 \\ 
        \hline
        Web & \href{https://zenodo.org/records/4656664}{Web Traffic} & ~ & ~ & ~ & ~ & ~ & ~ & ~ & ~ & ~ & ~ & ~ & ~ & ~ & ~ & ~ & ~ & ~ & ~ & ~ & ~ & ~ & ~ & ~ & $\surd$ & 1 \\ 
        \hline
        Energy & \href{https://zenodo.org/records/4656091}{London Smart Meters} & ~ & ~ & ~ & ~ & ~ & ~ & ~ & ~ & ~ & ~ & ~ & ~ & ~ & ~ & ~ & ~ & ~ & ~ & ~ & ~ & ~ & ~ & ~ & $\surd$ & 1 \\ 
        \hline
        Transport & \href{https://zenodo.org/records/5122232}{Rideshare} & ~ & ~ & ~ & ~ & ~ & ~ & ~ & ~ & ~ & ~ & ~ & ~ & ~ & ~ & ~ & ~ & ~ & ~ & ~ & ~ & ~ & ~ & ~ & $\surd$ & 1 \\ 
        \hline
        Transport & \href{https://zenodo.org/records/5122537}{Vehicle Trips} & ~ & ~ & ~ & ~ & ~ & ~ & ~ & ~ & ~ & ~ & ~ & ~ & ~ & ~ & ~ & ~ & ~ & ~ & ~ & ~ & ~ & ~ & ~ & $\surd$ & 1 \\ 
        \hline
        Nature & \href{https://zenodo.org/records/5129091}{Temperature Rain} & ~ & ~ & ~ & ~ & ~ & ~ & ~ & ~ & ~ & ~ & ~ & ~ & ~ & ~ & ~ & ~ & ~ & ~ & ~ & ~ & ~ & ~ & ~ & $\surd$ & 1 \\ 
        \hline
        Nature & \href{https://zenodo.org/records/4654822}{Weather\_AU} & ~ & ~ & ~ & ~ & ~ & ~ & ~ & ~ & ~ & ~ & ~ & ~ & ~ & ~ & ~ & ~ & ~ & ~ & ~ & ~ & ~ & ~ & ~ & $\surd$ & 1 \\ 
        \hline
        Multiple & \href{https://zenodo.org/records/4656548}{M4 Daily} & ~ & ~ & ~ & ~ & ~ & ~ & ~ & ~ & ~ & ~ & ~ & ~ & ~ & ~ & ~ & ~ & ~ & ~ & ~ & ~ & ~ & ~ & ~ & $\surd$ & 1 \\ 
        \hline
        Multiple & \href{https://zenodo.org/records/4656480}{M4 Monthly} & ~ & ~ & ~ & ~ & ~ & ~ & ~ & ~ & ~ & ~ & ~ & ~ & ~ & ~ & ~ & ~ & ~ & ~ & ~ & ~ & ~ & ~ & ~ & $\surd$ & 1 \\ 
        \hline
        Multiple & \href{https://zenodo.org/records/4656410}{M4 Quarterly} & ~ & ~ & ~ & ~ & ~ & ~ & ~ & ~ & ~ & ~ & ~ & ~ & ~ & ~ & ~ & ~ & ~ & ~ & ~ & ~ & ~ & ~ & ~ & $\surd$ & 1 \\ 
        \hline
        Multiple & \href{https://zenodo.org/records/4656379}{M4 Yearly} & ~ & ~ & ~ & ~ & ~ & ~ & ~ & ~ & ~ & ~ & ~ & ~ & ~ & ~ & ~ & ~ & ~ & ~ & ~ & ~ & ~ & ~ & ~ & $\surd$ & 1 \\ 
        \hline
        Energy & \href{https://zenodo.org/records/4654858}{Wind\_Farms\_AU} & ~ & ~ & ~ & ~ & ~ & ~ & ~ & ~ & ~ & ~ & ~ & ~ & ~ & ~ & ~ & ~ & ~ & ~ & ~ & ~ & ~ & ~ & ~ & $\surd$ & 1 \\ 
        \hline
        Transport & Taxi & ~ & ~ & ~ & ~ & ~ & ~ & ~ & ~ & ~ & ~ & ~ & ~ & ~ & ~ & ~ & ~ & ~ & ~ & ~ & ~ & ~ & ~ & ~ & $\surd$ & 1 \\ 
        \hline
        Transport & Uber & ~ & ~ & ~ & ~ & ~ & ~ & ~ & ~ & ~ & ~ & ~ & ~ & ~ & ~ & ~ & ~ & ~ & ~ & ~ & ~ & ~ & ~ & ~ & $\surd$ & 1 \\ 
        \hline
        Web & Wikipedia & ~ & ~ & ~ & ~ & ~ & ~ & ~ & ~ & ~ & ~ & ~ & ~ & ~ & ~ & ~ & ~ & ~ & ~ & ~ & ~ & ~ & ~ & ~ & $\surd$ & 1 \\ 
        \hline
        Transport & Air Passengers & ~ & ~ & ~ & ~ & ~ & ~ & ~ & ~ & ~ & ~ & ~ & ~ & ~ & ~ & ~ & ~ & ~ & ~ & ~ & ~ & ~ & ~ & ~ & $\surd$ & 1 \\ 
        \midrule[1.2pt]
        \multicolumn{2}{cI}{Total} & 7 & 5 & 9 & 6 & 5 & 6 & 9 & 3 & 6 & 5 & 6 & 6 & 5 & 5 & 6 & 4 & 9 & 5 & 4 & 5 & 3 & 8 & 20 & 26 & \\ 
        \bottomrule[2pt]
    \end{tabular}
         }
\end{table*}
\section{Input Data: How are datasets preprocessed before being fed into models?}
\label{datasets}
\subsection{Dataset Preparation}
We gather datasets from typical models, utilizing open-access papers and open-source codes, resulting in 50 datasets across 24 models~\footnote{We include the source of the code and link of dataset in the Tab.~\ref{Tran_LM}.}. These include 20 transformer models and 4 Large Language Models (LLMs) from 2020 to 2023. The transformer, an encoder-decoder architecture, is applied to various problems~\cite{islam2023comprehensive} such as NLP tasks, CV, and audio/speech processing. LLMs, which utilize transformer architecture, are pre-trained on large text datasets for NLP tasks. A recent survey~\cite{jin2023large} explores LLMs for time series data, making them relevant for comparison. The datasets and related models are shown in Tab.~\ref{Tran_LM}, which indicates that LLM-based time series models use more datasets than transformer-based time series models. These datasets do not have missing or corrupted data. Typically, missing data can be addressed through imputation, and corrupted data can be detected using anomaly detection algorithms. Most datasets for transformers-based TSF models are split into training, validation, and test sets, commonly at a 70/20/10 ratio, although this can vary. Papers~\cite{joseph2022optimal, muraina2022ideal} discuss optimal data splitting ratios, but these are not well-explored for transformer-based TSF models.
\subsection{Data Preprocessing}

Once data preparation has been completed, the next stage is data preprocessing. In this section, we follow the sequence of steps from the raw dataset to the model input. These steps are crucial for model performance to avoid ``Rubbish in, rubbish out''. The preprocessed data (model input) is then passed to the model~\cite{auffarth2021machine}. In transformer-based time series models, data undergoes sequential preprocessing: organizing features, data reduction or augmentation, and data representation for the model. (N.B. Data representation (input embedding and position encoding) is part of the data-model interaction, discussed in the Sec.~\ref{interaction}.)

\subsubsection{Data Features}
Data Features includes \emph{feature transformations} and \emph{feature engineering}. 
\emph{Feature transformations} transform a dataset into new distribution base on model's assumption. \emph{Feature Engineering} extracts features from input datasets to improve the performance of the models. A common feature transformation is data normalization, which is frequently used in transformer-based time series forecasting (TSF) models. Models apply data normalization to adjust data to the same common scale or range. This keeps different datasets and models on the same level for comparison. There are several normalization solutions such as Z-normalization, Min-max normalization, Sigmoid normalization etc.~\cite{lima2023large}. However, further research is needed to analyze the different data normalization methods used in transformer-based time series models. \emph{Feature Engineering} allows us to understand different features of time series data, which is essential for the performance of transformer-based forecasting models. Different features are discussed in~\cite{fulcher2018feature,hyndman2018forecasting,lazzeri2020machine}. Here, we focus on features applied in transformer-based time series forecasting. Below, we list some common feature engineering methods for transformer-based TSF models specifically.

\textbullet\texttt{Covariate:} 
In Dart~\cite{herzen2022darts}, covariate time series refer to external data or variables that are not the target of forecasting but are useful for improving forecasting accuracy. For example, when modeling participants' heart rates using their weight, additional factors such as environmental temperature and measurement time also influence heart rates. These factors are called covariates. The meaning of covariates can vary depending on the context. In some contexts, input features or explanatory variables are considered covariates~\cite{elsayed2021we}, similar to their definition in statistical dictionaries~\cite{everitt2010cambridge}. The paper by Davies~\cite{davies2024evaluating} discusses the role of covariates in forecasting models. The Temporal Fusion Transformer (TFT) model~\cite{wu2020adversarial} employs static covariate encoders to integrate covariates into the model.

\textbullet\texttt{Lag Features and Sliding/Rolling Window:} 
\texttt{Time lag} refers to previous steps in the time series. An example of lag is shown in Fig.\ref{rolling}. Lag-Llama\cite{rasul2023lag} apply time lags as covariates to build the forecasting model. The \texttt{Rolling Window} technique involves moving a window of specified length across the data sequentially. In traditional forecasting methods, statistics such as mean, median, and maximum are computed over a fixed-size sliding window. Fig.~\ref{rolling} illustrates this approach. Transformer-based forecasting models like Informer~\cite{zhou2021informer} utilize the sliding window method to construct the input dataset. An example demonstrating the use of a sliding window for encoder and decoder inputs is depicted in Fig.~\ref{slidingwindow}.
\begin{figure}[!ht]
\begin{center}
\begin{tikzpicture}[scale=0.9] 
\draw[thick, -Triangle] (0,0) -- (0.5\textwidth,0) node[font=\footnotesize,above left=10pt and -8pt]{Timestamp};
\foreach \x in {0,1,...,6}
\draw (\x cm,3pt) -- (\x cm,-3pt);
\foreach \x/\descr in {0,1,2,3,4,5,6}
\node[font=\scriptsize, text height=1.75ex,
text depth=.5ex] at (\x,-.3) {$\descr$};
\draw [thick,decorate,decoration={brace,amplitude=5pt}] (0,0.3)  -- +(6,0) 
       node [black,midway,above=4pt, font=\scriptsize] {Total Time};
\draw [thick,decorate,decoration={brace,amplitude=5pt}] (6,-.5) -- +(-5,0)
       node [black,midway,below=2pt, font=\scriptsize] {Lag 1};
\draw [thick,decorate,decoration={brace,amplitude=5pt}] (6,-1) -- +(-4,0)
       node [black,midway,below=2pt, font=\scriptsize] {Lag 2};
\draw [thick,decorate,decoration={brace,amplitude=5pt}] (6,-1.5) -- +(-3,0)
       node [black,midway,below=8pt, font=\scriptsize] {......};
\end{tikzpicture}
\hspace{0.1cm}
\begin{tikzpicture}[scale=0.9] 
\draw[thick, -Triangle] (0,0) -- (0.5\textwidth,0) node[font=\footnotesize,above left=10pt and -8pt]{Timestamp};
\foreach \x in {0,1,...,6}
\draw (\x cm,3pt) -- (\x cm,-3pt);
\foreach \x/\descr in {0,1,2,3,4,5,6}
\node[font=\scriptsize, text height=1.75ex,
text depth=.5ex] at (\x,-.3) {$\descr$};
\draw [thick,decorate,decoration={brace,amplitude=5pt}] (0,0.3)  -- +(6,0) 
       node [black,midway,above=4pt, font=\scriptsize] {Total Time};
\draw [thick,decorate,decoration={brace,amplitude=5pt}] (3,-.5) -- +(-3,0)
       node [black,midway,below=2pt, font=\scriptsize] {Window 1};
\draw [thick,decorate,decoration={brace,amplitude=5pt}] (4,-1) -- +(-3,0)
       node [black,midway,below=2pt, font=\scriptsize] {Window 2};
\draw [thick,decorate,decoration={brace,amplitude=5pt}] (5,-1.5) -- +(-3,0)
       node [black,midway,below=8pt, font=\scriptsize] {......};
\end{tikzpicture}
\caption{Left: Example of the Lag in Time Series. Right: Example of the Rolling Window (window size=3).}
\label{rolling}
\end{center}
\end{figure}
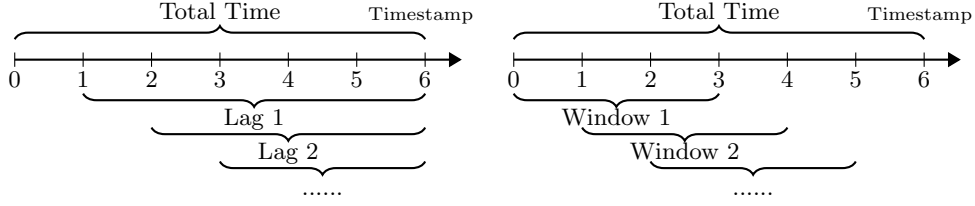
\begin{figure}[!ht]
\begin{center}
\includegraphics[width=0.8\linewidth]{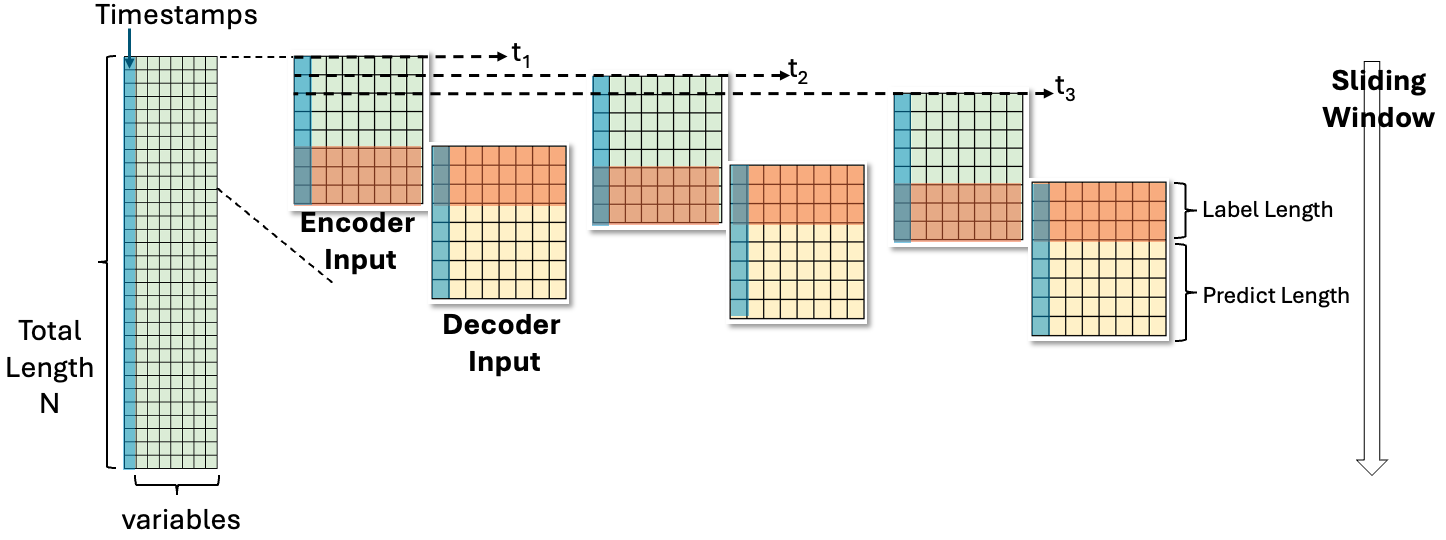}
\end{center}
\caption{Using Sliding Window to Prepare Encoder Input and Decoder Input for Transformer-based Time Series Models.}
\label{slidingwindow}
\end{figure}

\textbullet\texttt{Seasonal Decomposition:} 
Time series can generally be decomposed into three main components: Trend, Seasonal pattern, and Residual part~\cite{hartmann2023statistics}. The Trend component captures long-term increasing or decreasing patterns in the data. Seasonal pattern reflects periodic influences such as those caused by seasonal variations. The Residual component represents the remaining noise after extracting the Trend and Seasonal patterns. Models like FEDformer~\cite{zhou2022fedformer} and TDformer~\cite{zhang2022first} leverage this decomposition approach in their modeling strategies.

\textbullet\texttt{Stationarity and non-stationarity:} Formally, let $X_{t}=\{x_{t_{1}}, \ldots, x_{t_{n}}\}$ denote a time series, where $D(X_{t})$ represents its distribution function. $\tau$ denotes the time interval. A time series $X_{t}$ is stationary if $D(X_{t+\tau})$ does not depend on the observation time $t$. It means that time series behaves stochastically at any point in time. In contrast, non-stationary time series are influenced by trends or seasonality, exhibiting predictable patterns. Non-stationary Transformers~\cite{liu2022non} incorporate the concept of non-stationarity in data to inform their modeling approach.

\textbullet\texttt{Date Related:} 
In time series analysis, timestamps can range from seconds and minutes to hours, or calendar-based intervals such as days, weeks, and months. Each interval can reveal distinct patterns within the time series data. Moreover, factors like seasonality, holidays, and weekends introduce cyclical patterns. Date-related features serve as covariates in modeling these patterns.\\

\subsubsection{Data Reduction \& Data Augmentation}
\emph{Data Reduction} involves transforming the original data into a corrected and simplified form, often by cleaning up invalid data or generating summaries of the original dataset~\cite{turner1998travel}. Building upon the idea of summarization, PCATransformer~\cite{xu2023transformer} introduce Principal Component Analysis (PCA) based transformer TSF models, marking the initial exploration of data reduction in these models. However, research in this area remains limited. 
In contrast, \emph{Data Augmentation} is a well-explored technique used to enhance the size and quality of training datasets, thereby reducing overfitting in deep learning models such as transformers~\cite{shorten2019survey}. The survey~\cite{wen2021time} reviews various data augmentation methods for different time series tasks to summarize the augmentation methods.



\section{Data-Model Interaction: How does the data and model interact with each other?}
\label{interaction}
In this section, our aim is to address the interaction between data and the model. Specifically, we will focus on answering the question: \textit{How was the data prepared for the model, particularly for the encoder and decoder components?} In transformer-based forecasting architectures, the encoder-decoder framework takes a given time series as input and produces a predicted time series as output. While the general flow of data through a transformer model is described in the paper~\cite{ivanov2021data}, emphasizing hardware-level data movement during training, our paper primarily focuses on the data representation process within the transformer model. 
\begin{figure*}[!ht]
\begin{center}
\includegraphics[width=0.9\linewidth]{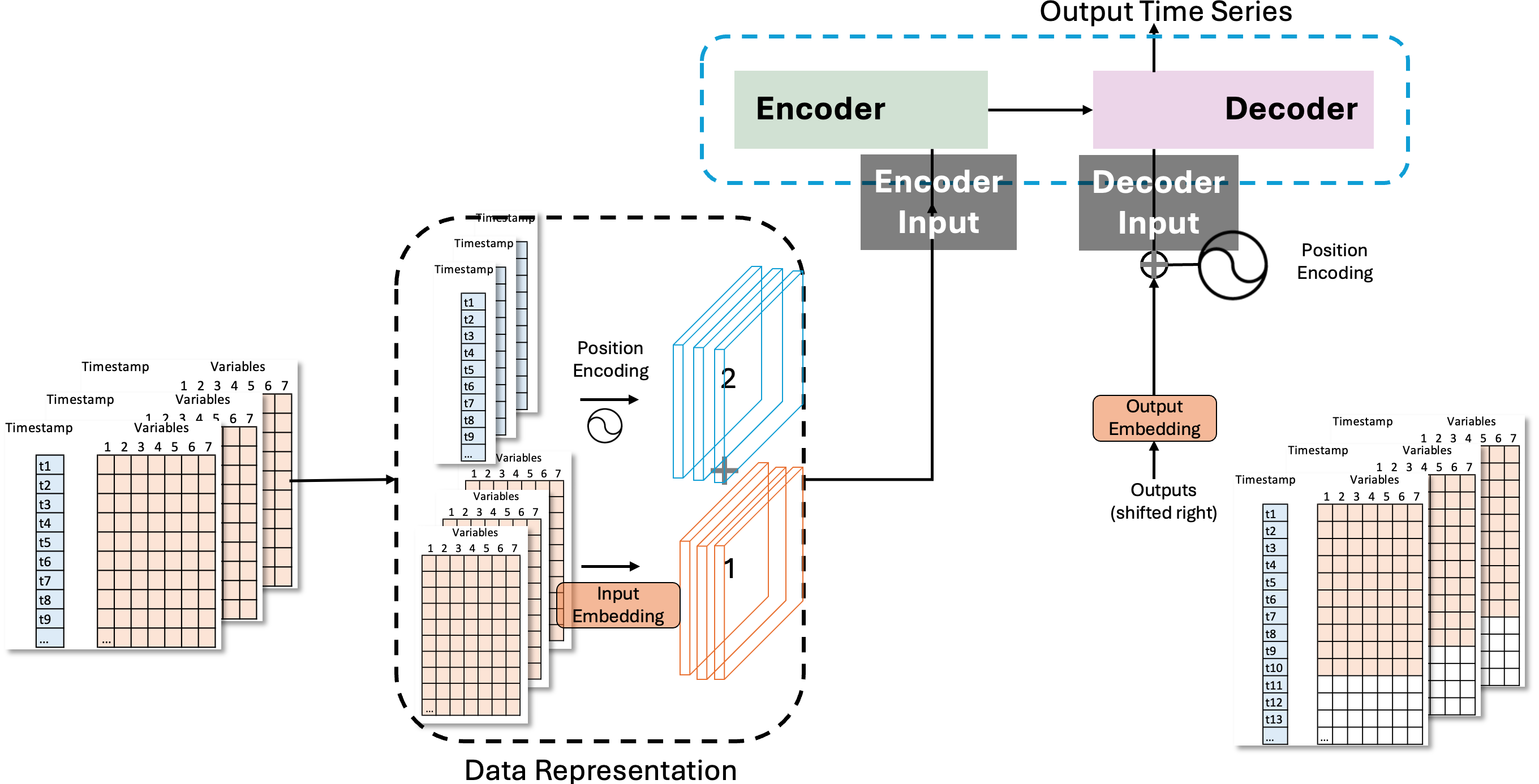}
\end{center} 
\caption{Input Embedding and Time Position Encoding}
\label{embedding}
\end{figure*}
\subsection{Data Representation: Input Embedding \& Time Position Encoding}
Fig.~\ref{embedding} illustrates the input embedding and time position encoding components in the transformer-based time series model. The inputs shown in the figure are preprocessed data prepared using sliding windows from Fig.~\ref{slidingwindow}. This preprocessed data includes a matrix of variables (features) and their corresponding timestamps. The variables (features) are represented using input embedding (the output is ``1'' in the Fig.~\ref{embedding}), while the timestamps are represented using time position encoding (the output is ``2'' in the Fig.~\ref{embedding}). Both representation processes result in matrices, which are then merged to form the input for the encoder. The input for the decoder is generated in the same manner. Input embedding techniques commonly include 1-D convolutional filters (with a kernel width of 3) on the input data, as used in models like Informer~\cite{zhou2021informer} and Autoformer~\cite{wu2021autoformer}, and linear projection methods, as seen in PatchTST~\cite{Yuqietal-2023-PatchTST}. Time position encoding typically applies temporal position encoding to represent timestamp information. The survey~\cite{dufter2022position} provides a comprehensive overview of existing position information methods in Transformer models, while the paper\cite{foumani2024improving} discuss various position encoding solutions for transformer-based time series models. The common sinusoidal position encoding is presented as follows, where $t$ is the $t-th$ timestamp, $j$ is the $j-th$ dimension of the model, $d$ is the total dimension of the model. 
\begin{equation}
    P_{tj}=
    \begin{cases}
    \centering
    sin(10000^{-\tfrac{j}{d}}t), & j\in {2n: n\in \mathbb{Z}}\\
    cos(10000^{-\tfrac{j-1}{d}}t), & j\in {2n+1: n\in \mathbb{Z}}
    \end{cases}
\end{equation}

\subsection{Modeling} 
\begin{figure}[!ht]
\begin{center}
\includegraphics[width=0.35\linewidth]{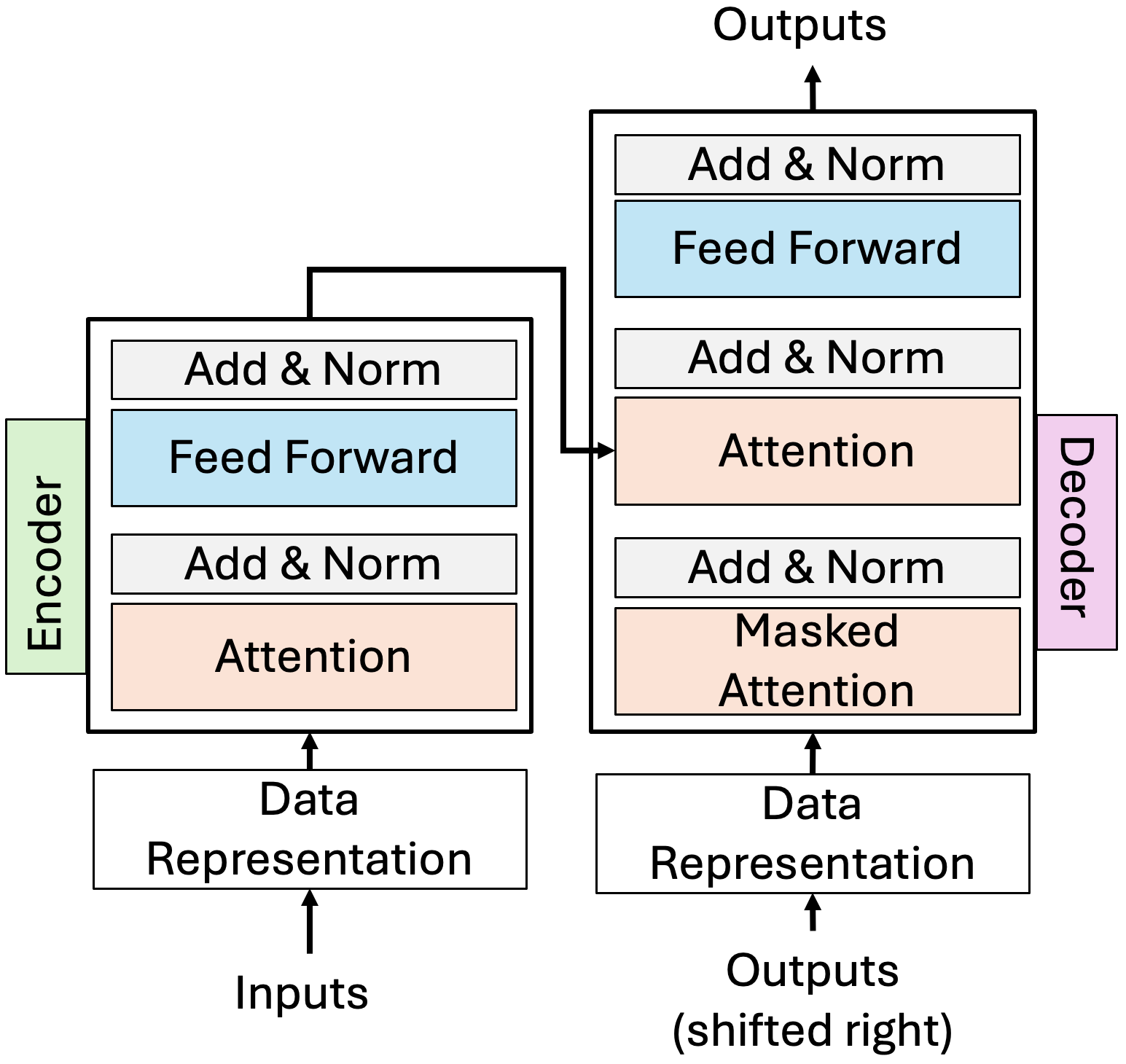}
\end{center} 
\caption{Main Components (Attention, Add\&Norm, Feed Forward) of the Transformer Model}
\label{maincomponents_trans}
\end{figure}

After data representation, the resulting matrices are fed into the model's first layer, known as the attention layer. This marks the end of the data-model interaction. The main components of the transformer model (Attention, Add \& Norm, and Feed Forward) are illustrated in Fig.~\ref{maincomponents_trans}. In the context of MCAI, transformer-based time series models are classified into different types based on modifications to these components and the overall architecture~\cite{liu2023itransformer, xu2023transformer}.

\section{Output Evaluation: How are models evaluated on the data?}
\label{evaluation}
Transformer-based time series models are generally evaluated from two perspectives: predictive performance and computational performance. Predictive performance is represented by metrics such as Mean Squared Error (MSE) and Mean Absolute Error (MAE). Computational performance is quantified by memory usage and computation time. These aspects are summarized in Table \ref{evaluation_table}. In this part, we summarize metrics used in transformer-based forecasting models and we discuss the computational performance measurement solution. 
\begin{table*}[!ht]
\begin{center}
\caption{Predictive Performance metrics and efficiency measurement solutions of transformer-based time series models}
\label{evaluation_table}
\resizebox{0.7\linewidth}{!}{
\begin{tabular}{ccc}
\toprule
\textbf{Model Abbrev.}  & \textbf{Predictive Performance}  & \textbf{Efficiency Measurements}   \\
\midrule
{GCformer}                    & MSE, MAE                  & training speed                       \\
{PDTrans}                     & Quantile Loss &                                      \\
{PatchTST}                    & MSE, MAE                  & runtime                                     \\
{Crossformer}                 & MSE, MAE                  & memeory, runtime              \\
{Scaleformer}                 & MSE, MAE                  & memeory, runtime                     \\
{Preformer}                   & MSE, MAE                  & memeory, runtime                     \\
{Client}                      & MSE, MAE                  & memeory, runtime, parameter quantity \\
{Taylorformer}                & Likelihood, MSE      &                                      \\
{iTransformer}                & MSE, MAE                  & memory                               \\
{TDformer}                    & MSE, MAE                  &                                      \\
{Non-stationary Trans.} & MSE, MAE                  &                                      \\
{FEDformer}                   & MSE, MAE                  &                                      \\
{TACTiS}                      & CRPS, Energy Score         &                                      \\
{Pyraformer}                  & MSE, MAE                  & memeory, runtime, Q-K pairs          \\
{TCCT}                        & MSE, MAE                  & memeory, runtime                     \\
{Triformer}                   & MSE, MAE                  & memeory                              \\
{Autoformer}                  & MSE, MAE                  & memeory, runtime                     \\
{Informer}                    & MSE, MAE                  & runtime                              \\
{TST}                         & Quantile Loss &                                      \\
{AST}                         & Quantile Loss &          \\                           
\bottomrule
\end{tabular}
}
\end{center}
\end{table*}
\subsection{Predictive Performance}
\subsubsection{Mean Squared Error (MSE) and Mean Absolute Error (MAE)} MSE and MAE are used to measure the difference between predicted values and actual values. Lower values of MSE and MAE indicate higher accuracy of the model. Considering $\hat{y_i}$ is predicted values and $y_i$ is the ground truth. The MSE and MAE is denoted as:
\begin{equation}
\begin{split}
    MSE = \frac{1}{n}\sum_{i=1}^{n}(\hat{y_i}-y_i)^2, 
    MAE = \frac{1}{n}\sum_{i=1}^{n}|\hat{y_i}-y_i| 
\end{split}
\end{equation}
\subsubsection{Quantile Loss ($\rho$-risk)}Quantiles divide a dataset into equal parts. Quantile is defined as $Q(\rho)=inf\{x: F(x) \ge \rho\}, 0<\rho<1$, where $F(x)$ is the distribution function~\cite{hyndman1996sample}. Normalized Quantile Loss, commonly used in transformer-based models, is discussed in~\cite{rangapuram2018deep, salinas2020deepar}. The $\rho$-quantile loss for $\rho \in (0,1)$ is defined as follow, where $\hat{x_i}$ is predicted values and $x_i$ is the ground truth. In the equation, if $x\ge\hat{x}$, then $P_{\rho}=\rho (x-\hat{x})$. This implies that larger values of $\rho$ assign more weight to the loss when $x\ge\hat{x}$.
\begin{equation}
\begin{split}
    L_{\rho}(x,\hat{x})=2\frac{\sum_{i=1}^{n}P_{\rho}(x_i,\hat{x_i})}{\sum_{i=1}^{n}x_i}, 
    P_{\rho}=
\begin{cases}
\rho (x-\hat{x}), x\ge\hat{x}\\
(1-\rho) (\hat{x}-x), x<\hat{x}
\end{cases}
\end{split}
\end{equation}

\subsubsection{Likelihood} The likelihood function is denoted as $L(\theta|X)=L(\theta|x_{1},...,x_{n})=f(x_{1},...,x_{n}|\theta)$ when ${x_{1},...,x_{n}}$ has density function $f(x_{1},...,x_{n}|\theta)$~\cite{lehmann2006theory}. Likelihood function $L(\theta|X)$ is the function of the parameters $\theta$ of the model. Taylorformer~\cite{nivron2023taylorformer}'s likelihood function is denoted as $L_{\theta}(Y_{T}|Y_{C},X_{C},X_{T})$, where $\{X_{C},Y_{C}\}$ is context set and $\{X_{T},Y_{T}\}$ is target set. It evaluates the model based on likelihood method and LocalTaylor function and neural network MHA-X-Net.

\subsubsection{Uncertainty Estimation} 
Unlike scoring function such as MSE and MAE, which measure the prediction quality, uncertainty estimation express measure the degree of the model's confidence in its own prediction. CRPS (Continuous Ranked Probability Score) is widely used, as it offer the  evaluation across the entire predictive distribution. The CRPS~\cite{grimit2006continuous,gneiting2007strictly,brocker2012evaluating} is defined as follows: 
\begin{equation}
\begin{split}
    CRPS(F,x)=\int (F(\hat{x})-H(\hat{x}-x))^2d\hat{x}, 
    H(\hat{x}-x)=
\begin{cases}
    0, x \ge \hat{x} \\
    1, x < \hat{x}
\end{cases}
\end{split}
\end{equation}

where $x \in \mathbb{R}$ is the observation, $F$ is the cumulative distribution function of the forecast distribution. As one variation of CRPS, TACTiS~\cite{drouin2022tactis} employs the CRPS calculation solution from GluonTS~\cite{gluonts_jmlr}, which exploits the fact that CRPS is equal to twice the mean quantile loss (as per~\cite{brocker2012evaluating}). Another related metric is the Energy Score~\cite{gneiting2007strictly} which characterizes the equality of distributions~\cite{szekely2003statistics}.

\subsection{Computational Performance}
Most papers record memory usage, runtime, or speed to measure a model's efficiency. Pyraformer~\cite{liu2021pyraformer} uses the number of query-key dot products (Q-K pairs) to describe the time and space complexity of the model. However, with the growing importance of sustainability in the face of climate change, it is crucial to develop models with lower environmental impact. Therefore, we also measure the carbon footprint of models (See Appendix.~\ref{secA1}) to raise awareness and encourage the development of sustainable models.

\section{Future Research Opportunities}
\label{challenges}
\subsection{Input Dataset}
The Monash Time Series Forecasting Archive~\cite{godahewa2021monash} offers a diverse range of comprehensive time-series datasets across various domains, accompanied by dataset characteristic analyses. However, there is still a lack of investigation of dataset exploration pipelines tailored specifically for transformer-based time series models. Additionally, research on determining optimal split ratios (training, validation, and testing) remains insufficient, with only a few papers such as~\cite{joseph2022optimal, muraina2022ideal} discussing the subject. Moreover, while researchers have begun focusing on dataset augmentation in transformer-based time series models~\cite{wen2021time}, dataset reduction for transformer models has so far been neglected. Dataset reduction is crucial due to the complexity (long timestamps with large multiple variables) of time series datasets. Furthermore, most transformer models primarily operate on common datasets such as electricity consumption, traffic, and exchange rates (as shown in Table~\ref{Tran_LM}). Some recent papers~\cite{wu2024strategic, wu2024trustworthy} have explored transformer-based time series models in the financial domain. However, further investigation into the application of transformer-based time series models on diverse real-world datasets is desirable. Real-world datasets are often a mix of various data types, such as medical records including patients' images or speech data alongside numerical measurements~\cite{waqas2023multimodal}. As researchers are also delving into spatio-temporal data~\cite{jin2023large, zhang2024large}, this signals a growing interest in transformer-based multimodal forecasting for using various data types.

\subsection{Input Data Representation}
In this paper, the input data representation includes the input embedding and time position encoding. Another survey paper~\cite{dufter2022position} has explored position information methods in transformer models, highlighting the importance in capturing sequential relationships effectively. Recent research in particular~\cite{foumani2024improving} has studied these methods to suit temporal data under the context of transformer-based time series models. Despite advancements in position encoding techniques, research on optimizing input embedding strategies tailored for transformer-based time series models remains insufficient. 

\subsection{Evaluation}
Most transformer-based models use Mean Squared Error (MSE) and Mean Absolute Error (MAE) to assess forecast errors between predicted and observed values. Some models employ quantile loss to evaluate prediction intervals, while others utilize probability-based metrics such as likelihood and Continuous Ranked Probability Score (CRPS) to assess the alignment of predicted probability distributions with observed values. However, understanding the reliability and trustworthiness of these models remains challenging due to the opacity of neural networks within transformers. Methods for identifying, quantifying, and communicating uncertainties in model outputs are discussed in the book~\cite{loucks2017water}. Surveys~\cite{hu2023uncertainty} explore uncertainty management techniques in NLP from both data and model perspectives. However, uncertainty management in transformer-based time series haven't been investigated enough. A recent paper~\cite{sergazinov2023gluformer} quantified uncertainty in transformer-based TSF models using glucose datasets. Robformer~\cite{yu2024robformer} innovates with a robust decomposition module to address trend shifting. However, generalizing these results to other datasets and modules is decisive. 
\section{Conclusion}
\label{Conclusion}
This paper has explored the role of data-centric AI in transformer-based time series forecasting by addressing three key research questions and proposing a taxonomy. Firstly, in the Input Data section, we addressed RQ1 \textit{How datasets are preprocessed before being fed into models?} by discussing the data preparation and preprocessing in transformer-based time series forecasting. Secondly, in the Data-Model Interaction section, we answered RQ2 \textit{How does the data and model interact with each other?} by delving into the data representation within transformer-based time series forecasting models. Finally, in the Output Evaluation section, we addressed RQ3 \textit{How are models evaluated based on the data?} Furthermore, we highlight future research opportunities based on these three reseach questions.
\bibliography{sn-bibliography}
\newpage
\begin{appendices}
\section{Experiments on $CO_2$ emission}
\label{secA1}
The experiments on $CO_2$ emissions were conducted on the High Performance Computing (HPC) using commonly utilized datasets as listed in Tab.~\ref{ds_bench} (Also see Tab.~\ref{Tran_LM}). On HPC, the GPU setting is \{Tesla V100-SXM2-16GB and Tesla V100-SXM2-32GB\}. The CPU setting is \{Intel(R) Xeon(R) Gold 6132 CPU @ 2.60GHz/2 device(s), TDP:140.0\}. The transformer-based forecasting models and their settings follow the Time-Series-Library, with the measurement solution provided by ECO2AI~\cite{budennyy2023eco2ai}. This experiment was conducted in a long-term forecasting setting with different prediction lengths. 

\begin{figure}[!ht]
\begin{center}
\includegraphics[width=\linewidth]{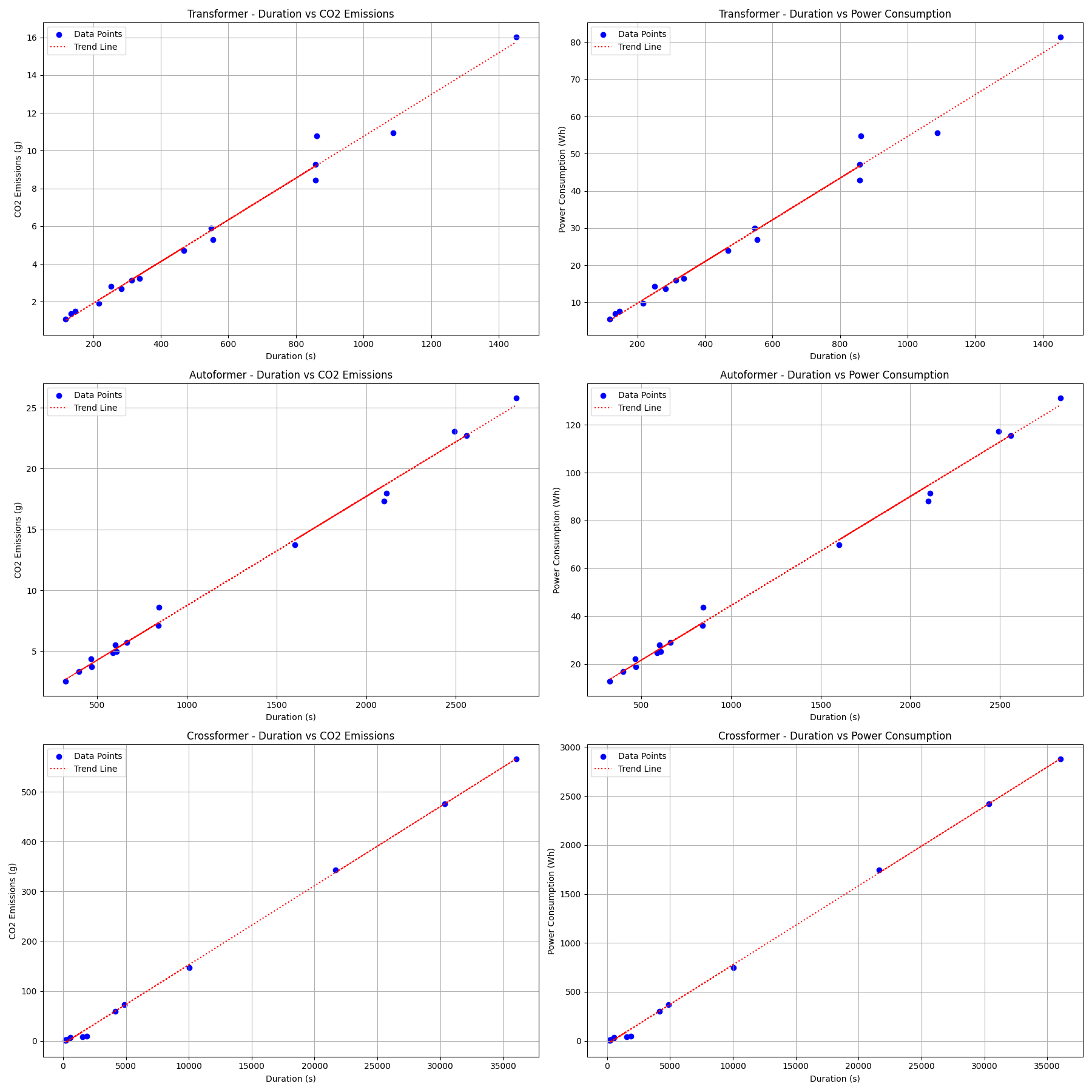}
\end{center}
\caption{Scatter Trend Charts of $CO_2$ Emission and Runtime (duration)/Power Consumption on Transformer-based Long-term Forecasting Models.}
\label{ScatterTrendCharts}
\end{figure}

\begin{figure}[!ht]
\begin{center}
\includegraphics[width=\linewidth]{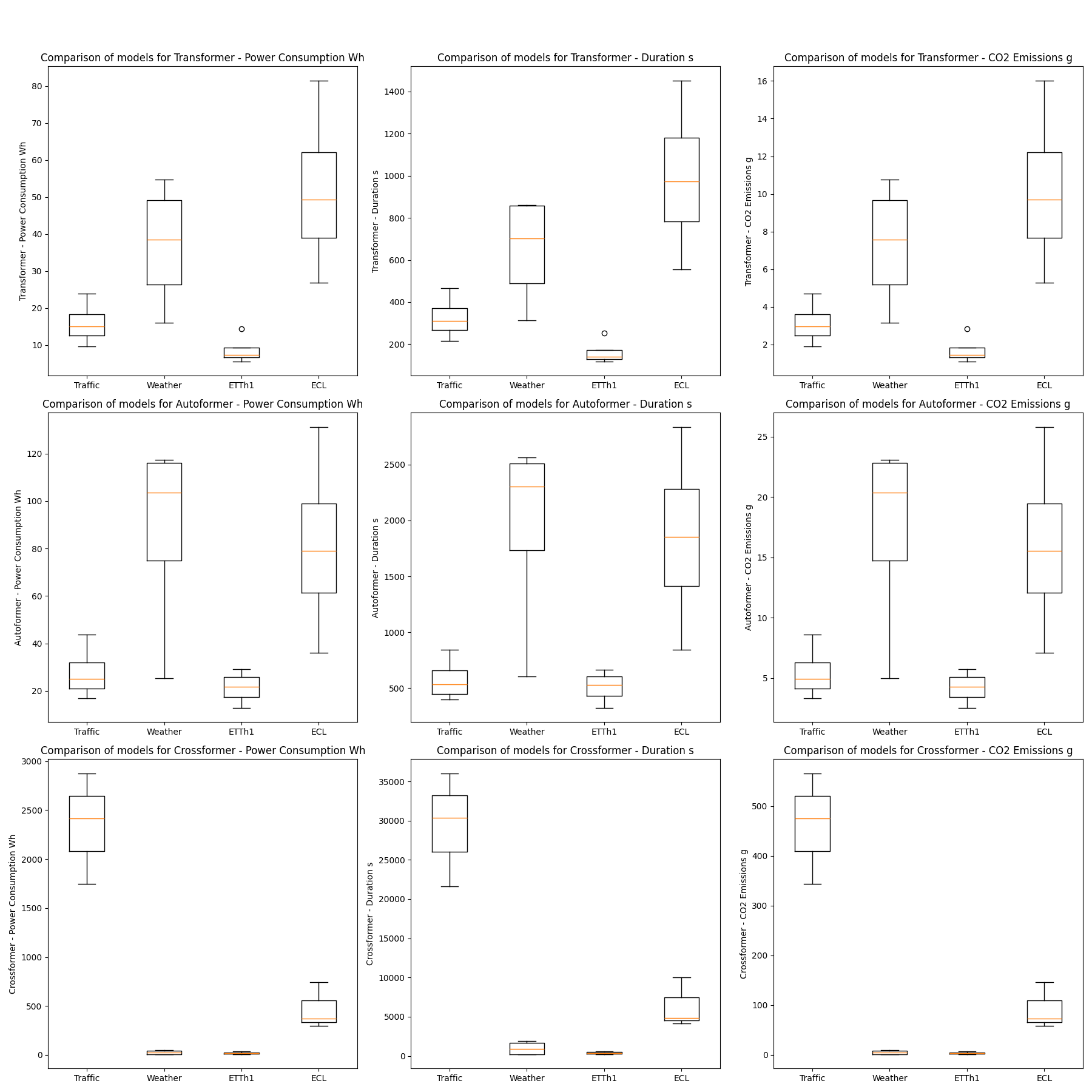}
\end{center}
\caption{Box Plot of $CO_2$ Emission of Different Models on Different Datasets.}
\label{BoxPlot}
\end{figure}

In Table~\ref{tab:Training_results}, the settings \{96, 192, 336, 720\} correspond to the prediction lengths. The results indicate that longer training times lead to higher CO2 emissions(See Fig.~\ref{ScatterTrendCharts}). Notably, the Crossformer model exhibited the highest CO2 emission at 566.09 grams, equivalent to burning 250 milliliters of gasoline~\footnote{see calculation at~\href{https://natural-resources.canada.ca/sites/www.nrcan.gc.ca/files/oee/pdf/transportation/fuel-efficient-technologies/autosmart_factsheet_6_e.pdf}{link}}. In addition, The $CO_2$ emission is influenced by training dataset, especially by the number of timestemps and the number of variables. In Fig.~\ref{BoxPlot}, Transformer and Autoformer are influenced by the number of timestamps (The more timestemps, the more $CO_2$ emission. See Weather and ECL dataset on Transformer and Autoformer model). Crossformer is influenced by the number of variables (The more variables, the more $CO_2$ emission. See Traffic and ECL dataset on Crossformer model). 
\begin{table}[!ht]
    \centering
    \caption{Summary of information about four datasets used in measuring $CO_{2}$ emission).}
    \label{ds_bench}
    \begin{tabular}{c|c|c|c|c}
     \toprule
        Datasets & ETTh1 & Weather & Electricity (ECL) & Traffic \\ 
        \hline
        Variables & 7 & 21 & 321 & 862 \\ 
        \hline
        Timestamps & 17420 & 52696 & 26304 & 17544 \\
    \bottomrule
    \end{tabular}
\end{table}

\begin{table*}[!ht]
\caption{$CO_{2}$ Emissions of Training Process in Transformer-based Time Series Forecasting (OOM: Out of Memory)}
\label{tab:Training_results}
\resizebox{\linewidth}{!}{
\begin{tabular}{cc|ccc|ccc|ccc}
\hline
\multicolumn{2}{c|}{Models}                         & \multicolumn{3}{c|}{Autoformer}                                                                                                                                                           & \multicolumn{3}{c|}{Crossformer}                                                                                                                                                          & \multicolumn{3}{c}{Transformer}                                                                                                                                                           \\ \hline 
\multicolumn{2}{c|}{Metric}                         & \begin{tabular}[c]{@{}c@{}}duration\\ (s)\end{tabular} & \begin{tabular}[c]{@{}c@{}}power\\consumption(Wh)\end{tabular} & \begin{tabular}[c]{@{}c@{}}CO2\\emissions(g)\end{tabular} & \begin{tabular}[c]{@{}c@{}}duration\\ (s)\end{tabular} & \begin{tabular}[c]{@{}c@{}}power\\consumption(Wh)\end{tabular} & \begin{tabular}[c]{@{}c@{}}CO2\\emissions(g)\end{tabular} & \begin{tabular}[c]{@{}c@{}}duration\\ (s)\end{tabular} & \begin{tabular}[c]{@{}c@{}}power\\consumption(Wh)\end{tabular} & \begin{tabular}[c]{@{}c@{}}CO2\\emissions(g)\end{tabular} \\ \hline
\multicolumn{1}{c|}{\multirow{4}{*}{ECL}}     & 96  & 841.8177                                               & 36.0394                                                           & 7.0897                                                       & 4157.4333                                              & 299.0629                                                          & 58.8320                                                      & 555.0109                                               & 26.8174                                                           & 5.2756                                                       \\
\multicolumn{1}{c|}{}                         & 192 & 2099.2409                                              & 88.1902                                                           & 17.3489                                                      & 4859.8069                                              & 369.5088                                                          & 72.6901                                                      & 857.2692                                               & 42.9433                                                           & 8.4478                                                       \\
\multicolumn{1}{c|}{}                         & 336 & 1602.8052                                              & 69.9166                                                           & 13.7541                                                      & 10024.7159                                             & 745.7526                                                          & 146.7052                                                     & 1088.4216                                              & 55.6925                                                           & 10.9559                                                      \\
\multicolumn{1}{c|}{}                         & 720 & 2837.3136                                              & 131.2501                                                          & 25.8196                                                      & OOM                                                     & OOM                                                                & OOM                                                           & 1452.2351                                              & 81.4772                                                           & 16.0283                                                      \\ \hline
\multicolumn{1}{c|}{\multirow{4}{*}{ETTh1}}   & 96  & 322.3691                                               & 12.7535                                                           & 2.5089                                                       & 217.5166                                               & 8.4103                                                            & 1.6545                                                       & 117.5711                                               & 5.5337                                                            & 1.0886                                                       \\
\multicolumn{1}{c|}{}                         & 192 & 468.7925                                               & 18.9128                                                           & 3.7205                                                       & 247.1342                                               & 10.7373                                                           & 2.1122                                                       & 134.2723                                               & 6.9900                                                            & 1.3751                                                       \\
\multicolumn{1}{c|}{}                         & 336 & 663.4532                                               & 29.1745                                                           & 5.7392                                                       & 492.4600                                               & 24.4043                                                           & 4.8008                                                       & 146.9931                                               & 7.5727                                                            & 1.4897                                                       \\
\multicolumn{1}{c|}{}                         & 720 & 587.0618                                               & 24.6777                                                           & 4.8546                                                       & 560.1151                                               & 34.2809                                                           & 6.7438                                                       & 251.7572                                               & 14.3321                                                           & 2.8194                                                       \\ \hline
\multicolumn{1}{c|}{\multirow{4}{*}{Traffic}} & 96  & 397.8709                                               & 16.9082                                                           & 3.3262                                                       & 21642.9455                                             & 1746.5806                                                         & 343.5891                                                     & 216.2519                                               & 9.7177                                                            & 1.9117                                                       \\
\multicolumn{1}{c|}{}                         & 192 & 466.2856                                               & 22.2510                                                           & 4.3772                                                       & 30353.6828                                             & 2418.3465                                                         & 475.7395                                                     & 283.9747                                               & 13.5821                                                           & 2.6719                                                       \\
\multicolumn{1}{c|}{}                         & 336 & 600.3932                                               & 28.0720                                                           & 5.5224                                                       & 36043.9070                                             & 2877.6318                                                         & 566.0906                                                     & 337.8724                                               & 16.3602                                                           & 3.2184                                                       \\
\multicolumn{1}{c|}{}                         & 720 & 844.6213                                               & 43.6340                                                           & 8.5837                                                       & OOM                                                     & OOM                                                                & OOM                                                           & 467.6036                                               & 23.9352                                                           & 4.7086                                                       \\ \hline
\multicolumn{1}{c|}{\multirow{4}{*}{Weather}} & 96  & 607.6077                                               & 25.2550                                                           & 4.9682                                                       & 1884.1530                                              & 46.8592                                                           & 9.2182                                                       & 313.3538                                               & 16.0077                                                           & 3.1490                                                       \\
\multicolumn{1}{c|}{}                         & 192 & 2111.8946                                              & 91.4837                                                           & 17.9968                                                      & 186.7157                                               & 5.0592                                                            & 0.9953                                                       & 858.2094                                               & 47.2053                                                           & 9.2863                                                       \\
\multicolumn{1}{c|}{}                         & 336 & 2560.8089                                              & 115.5892                                                          & 22.7388                                                      & 1544.5321                                              & 42.6614                                                           & 8.3924                                                       & 548.1598                                               & 29.8991                                                           & 5.8818                                                       \\
\multicolumn{1}{c|}{}                         & 720 & 2492.1999                                              & 117.2282                                                          & 23.0613                                                      & 232.2534                                               & 7.3457                                                            & 1.4451                                                       & 861.5856                                               & 54.7474                                                           & 10.7700                                                      \\ \hline
\end{tabular}}
\end{table*}

\end{appendices}

\end{document}